\documentclass[letterpaper, 10 pt, conference]{ieeeconf}  %

\IEEEoverridecommandlockouts                
\usepackage{amsmath}
\usepackage{amssymb}
\usepackage{amsfonts}

\usepackage{tabu}
\usepackage{booktabs} %
\usepackage{multirow}
\usepackage{adjustbox}

\usepackage{xcolor}
\usepackage{xspace}
\usepackage{ifthen}

\usepackage{graphicx}
\usepackage[font=small]{caption}
\usepackage{tikz}
\usetikzlibrary{positioning}

\let\labelindent\relax %
\usepackage{enumitem} %

\usepackage{autolabtools}

\newcommand{\ito}{\mathrel{{.}\,{.}}\nobreak}
\newcommand{\irange}[2]{[#1 \ito #2]}

\usepackage[backend=biber,
            hyperref=true,
            url=false,
            isbn=false,
            doi=false,
            backref=false,
            style=ieee,
            natbib=true,%
            mincitenames=1,
            maxcitenames=1,
            citestyle=numeric-comp,
            sorting=nyt,%
            block=none]{biblatex}
            
\usepackage{comment}
\title{\LARGE \bf
Mechanical Search on Shelves using a Novel ``Bluction'' Tool
}

\author{Huang Huang$^1$, Michael Danielczuk$^1$, Chung Min Kim$^1$, Letian Fu$^1$, Zachary Tam$^1$,\\
Jeffrey Ichnowski$^1$, Anelia Angelova$^2$, Brian Ichter$^2$, and Ken Goldberg    $^1$%

\thanks{
$^{1}$The AUTOLab at University of California, Berkeley.
$^{2}$Google Brain.}
}

\newcommand{\AlgName}{SLAX-RAY}
\newcommand{\bluc}{bluction\xspace}

\begin{document}
\maketitle
\begin{abstract}
Shelves are common in homes, warehouses, and commercial settings due to their storage efficiency. However, this efficiency comes at the cost of reduced visibility and accessibility. When looking from a side (lateral) view of a shelf, most objects will be fully occluded, resulting in a constrained lateral-access mechanical search problem. To address this problem, we introduce: (1) a novel \emph{bluction} tool, which combines a thin pushing blade and suction cup gripper, (2) an improved LAX-RAY simulation pipeline and perception model that combines ray-casting with 2D Minkowski sums to efficiently generate target occupancy distributions, and
(3) a novel SLAX-RAY search policy, which optimally reduces target object distribution support area using the bluction tool. Experimental data from 2000 simulated shelf trials and 18 trials with a physical Fetch robot equipped with the bluction tool suggest that using suction grasping actions improves the success rate over the highest performing push-only policy by 26\% in simulation and 67\% in physical environments.
\end{abstract}
\section{Introduction} \label{sec:introduction}
Finding a target object in lateral-access settings such as shelves---common in homes, warehouses, and retail stores---is complicated by objects toward the front blocking visibility of and access to objects further back.
In contrast with the overhead-access bin settings explored in prior works~\cite{danielczuk2019mechanical,kurenkov2020visuomotor,yang2020deep,zeng2018learning}, where objects can be heaped in arbitrary poses, objects on shelves consistently rest in stable poses.
If a target object is occluded on a shelf, other objects must be carefully moved without toppling to reveal the target.

Prior work on mechanical search in lateral-access environments has used pushing actions, executed with a wrist-mounted blade~\cite{huang2020mechanical}. This paper introduces a novel end-effector that combines a blade for pushing and a suction cup for grasping. The blade and suction cup are mounted on a thin shaft that can be rotated to maximize camera visibility. We call this end-effector the bluction tool (a pormanteau of ``blade'' and ``suction'').
\begin{figure}
    \centering
    \input{figures/bluction_diagram} \\[4pt] %
    \includegraphics[width=\columnwidth]{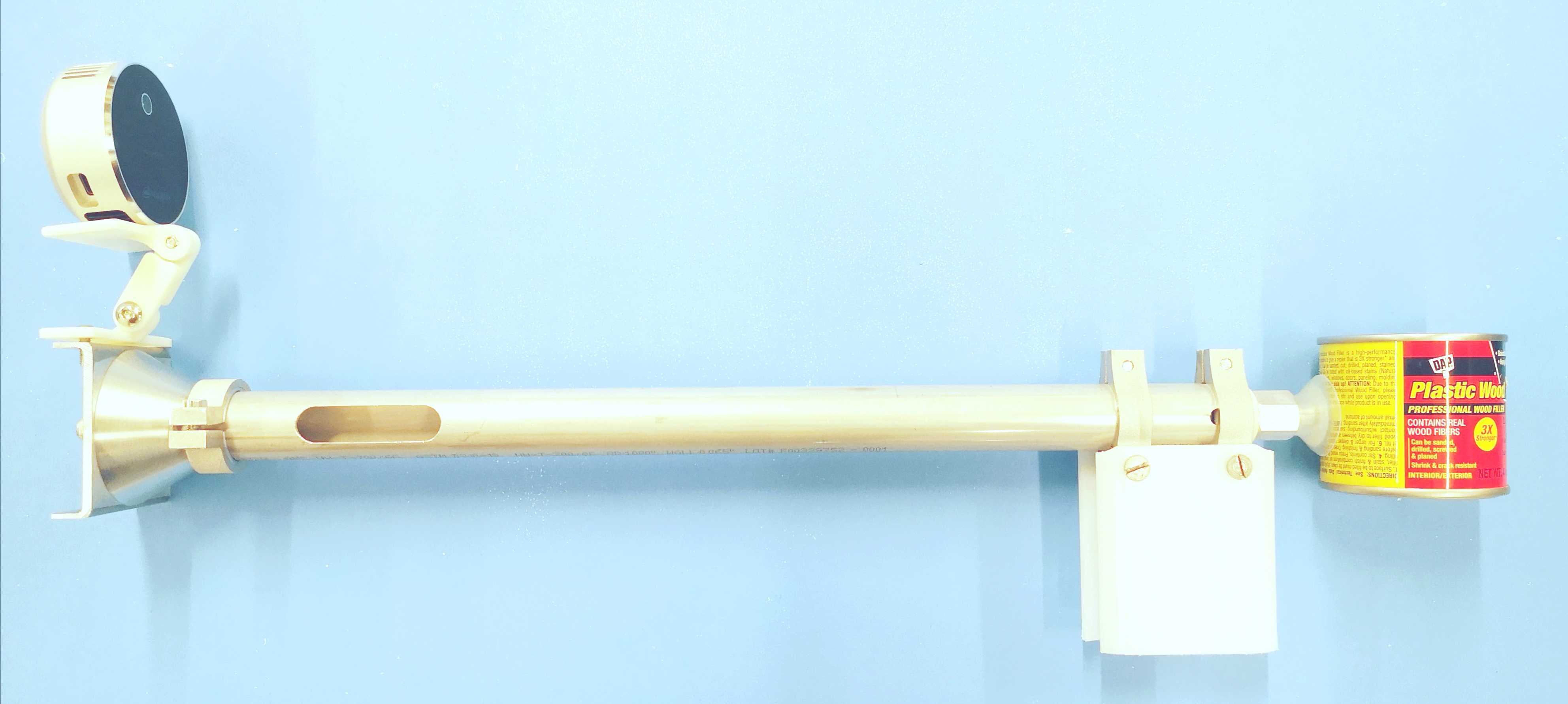} \\[4pt]
    \includegraphics[width=\linewidth]{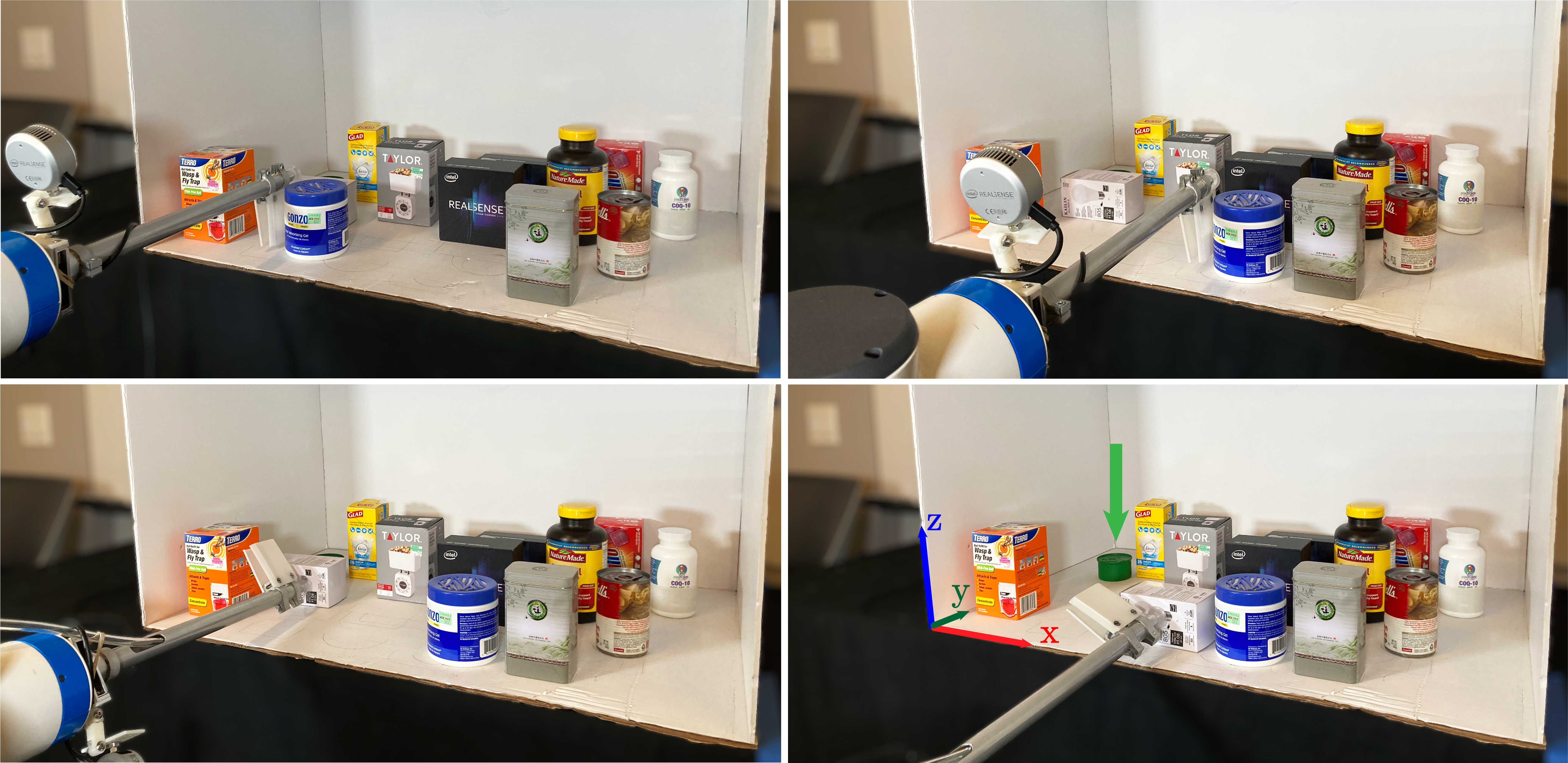}
    \caption{\textbf{Bluction Tool}: CAD (top) and Physical Implementation (middle). Using the bluction tool, the SLAX-RAY policy performs pushing and suction actions (bottom) to reveal the green cube target object among occluding objects within the shelf. Suction actions allow the robot to increase the efficiency of the search and reveal the target object in scenes where no pushing actions are available. The shelf coordinate system is shown in the bottom right.}
    \vspace{-4pt}
    \label{fig:bluction_splash}
\end{figure}

The expanded action set afforded by the bluction tool is particularly beneficial when objects are close together or close to the shelf wall. In prior work, these objects would likely be immovable as the robot would have no space to place its end-effector to reach and push them, and some pushes could result in unrecoverable failures.  Thus, even moderately dense scenes could make target extraction impossible. Suction grasping can resolve these issues, but requires new planning algorithms that aim to balance pushing and suction grasping actions to efficiently reveal the target object.

This paper introduces \emph{Suction Lateral Access maXimal Reduction in support Area of occupancY distribution} (SLAX-RAY), and makes the following contributions:
\begin{enumerate}
    \item The design and evaluation of a novel robot end-effector: the bluction tool.

    \item An improved LAX-RAY simulation pipeline and perception model, which uses ray-casting and 2D Minkowski sums to generate 100,000 random shelf configurations with a fully-occluded target. The perception model learns from these configurations and their corresponding occupancy distributions to estimate a 1D distribution for the position of the target object.
    \item A lateral-access mechanical search policy using the bluction tool: Bluction-DAR, which attempts to maximally reduce the support of the target occupancy distribution at each timestep through a combination of suction and pushing actions.
   
    \item Experimental data from 2000 simulated environments with 4 -- 10 occluding objects and 18 physical environments with a Fetch robot with 8 -- 12 occluding objects, that suggests the SLAX-RAY search policy can improve success rate by $26\%$ in simulation and $67\%$ in physical compared to pushing-only baseline policies.
\end{enumerate}

\section{Related Work} \label{sec:relatedwork}
\subsection{Mechanical Search}
For mechanical search for a specific target object in an overhead-access environment (e.g., a tabletop or bin), the robot uses an overhead camera to acquire color, depth, or RGBD images of the scene and iteratively uses top-down grasps, pushes, or a combination of these actions to reveal and extract the desired target object. \citet{danielczuk2019mechanical} formulate the mechanical search problem and introduce a two-stage perception and search policy pipeline that uses heuristic high-level policies to guide pushing and grasping within the bin. \citet{kurenkov2020visuomotor} extend this work by introducing a learned, non-linear pushing action to uncover the target. \citet{zeng2018learning}, \citet{novkovic2020object}, and \citet{yang2020deep} explore the problem in a tabletop setting and jointly learn coordinated pushing and grasping strategies. One promising approach for mechanical search is to statistically estimate target object locations. \citet{xiao2019online} and \citet{price2019inferring} attempt to model this distribution using particle filtering and shape completion approaches respectively. \citet{danielczuk2020x} explicitly learn the distribution in image space by generating a large dataset of simulated depth images and corresponding target object occupancy distributions. The target occupancy distributions represent all locations of the target object that would result in the rendered depth image. They then train a neural network to estimate this distribution for multiple target object aspect ratios and show it can transfer to physical scenes as part of an overhead-access mechanical search policy.

However, as noted in Section~\ref{sec:introduction}, the lateral-access shelf environment introduces new challenges and constraints. \citet{zeng2017multi} use pick-and-place actions in a shelf environment as part of the Amazon Picking Challenge, but the robot does not rearrange objects \textit{within} the shelf. \citet{motoda2021bimanual} consider bimanual suction grasping to extract a target object from a stack on a shelf by predicting which objects will collapse given the target being removed; however, they focus on a single-step case in which the target is visible, whereas we consider the multi-step case where the target object is initially occluded. \citet{li2016act} use a POMDP solver in sparse simulated shelf environments, and show that their approach can outperform a greedy approach. \citet{bejjani2020occlusion} learn a policy that pushes occluding objects aside as the robot reaches for the target using a learned distribution over likely target states in the shelf. \citet{gupta2013interactive} introduce a multi-step object search algorithm using a PR2 robot to push or pick-and-place objects in a shelf; in contrast, we leverage the learned target object occupancy distribution to inform the search and do not discretize the shelf during the search. \citet{huang2020mechanical} introduce LAX-RAY with two pushing policies for the shelf environment. The LAX-RAY policies leverage a history encoding of target occupancy distributions and multi-step lookahead to efficiently reveal a target object on a shelf with up to eight occluding objects. In this paper, we use introduce the \bluc tool for suction grasping actions in addition to pushing actions, and introduce a bluction-based policy that maximally reduces distribution area at each timestep.

\subsection{Suction Grasping}
Suction-based end-effectors are commonly used in industrial picking as they can grasp objects with a single point of contact. They have also been used in robotics contexts ranging from the Amazon Picking Challenge~\cite{zeng2017multi,morrison2018cartman} and grasping~\cite{mahler2018dex,mahler2019learning,valencia20173d} to underwater manipulation~\cite{stuart2015suction} and wall climbing~\cite{bahr1996design}. 

When planning suction grasps, existing approaches either directly propose grasps on a point cloud of the scene using heuristic methods or sample a range of candidate grasps and rank them using a quality metric. For the former, common approaches are grasping near estimated centroids of flat surfaces~\cite{yu2016summary}, grasping along inward surface normals towards an object centroid~\cite{hernandez2016team}, or pushing objects from the top or side until a suction seal is formed~\cite{eppner2016lessons}. In the latter case, \citet{domae2014fast} use a geometric model that assesses planarity by convolving a contact template with the image. Cartman~\cite{morrison2018cartman}, which won the 2017 Amazon Picking Challenge, ranked grasps according to their distance from object boundaries. Recently, \citet{zeng2017multi} and \citet{cao2021suctionnet} use large, labeled datasets of real images to train a neural network that predicts grasp affordances directly from RGBD images or point clouds. Similarly, \citet{mahler2018dex} use a hybrid approach wherein suction grasps are modeled in wrench space and labeled on simulated depth images. A learned network trained on these simulated data is applied to real depth images.

In the majority of these papers, grasping with suction end-effectors is considered in an overhead scenario where the suction grasp approach axis is antiparallel to gravity~\cite{mahler2019learning}. In those settings, the elasticity of the cup and the vacuum force must counteract gravity, but shear forces are rare. In this paper, we instead focus on suction grasps with an approach axis perpendicular to gravity, which results in higher shear forces that the suction cup must resist.

\section{Problem Statement} \label{sec:problemstatement}
A set of objects rest in stable poses within a shelf, with one object designated as the target to be revealed within the shelf. Only the target is of known color and geometry. The robot views the shelf from the open side using an RGBD camera that is attached to its arm.

We additionally assume:
\begin{itemize}
    \item The shelf and camera poses are static with respect to the robot.
    \item All objects are cylinders or rectangular prisms with a flat face resting on the shelf surface.
    \item Actions do not inadvertently topple objects or move multiple objects simultaneously.
\end{itemize}

At any time $t \in \{1, \ldots, H\}$, where $H$ is a user-specified time limit, let
$y_t \in \mathbb{R}^{w \times h \times 4}$ be the observation from an RGBD camera of 
$s_t \in \mathcal{S}$, the arrangement of objects on the shelf.
At time $t$, the robot performs an action $a_t \in \mathcal{A}$, where $\mathcal{A} = \mathcal{A}_p \cup \mathcal{A}_s$, the union of pushing and suction actions.  The search terminates in \emph{failure} after $H$ steps, and in \emph{success} when the robot observes a threshold visibility $v\%$ of the target object.

Pushing actions in $\mathcal{A}_p$, parametrized as $\mathbf{a}_t = (\mathbf{q}, d)$, start with the blade at $\mathbf{q} \in \mathbb{R}^3$ 
relative to the center of the shelf, and push a signed distance $d \in \mathbb{R}$ along the $x$-axis of the shelf frame, which is shown in Fig.~\ref{fig:bluction_splash}.

Suction actions in $\mathcal{A}_s$, parametrized as $\mathbf{a}_t = (\mathbf{q}, d_x, d_z)$, start with the robot forming 
a seal between the suction cup and object, followed by 4 linear motions:
(1) lifting the object slightly,
(2) pulling the object towards the camera,
(3) translating the object horizontally by $d_x$, and 
(4) pushing the object into its final placement pose at a distance $d_z \in \mathbb{R}$ into the shelf.

We also define an action cost ratio, $\psi > 1$, to account for the increased motion planning and execution time for suction actions as compared to pushing actions. By measuring action planning and execution times during physical experiments, we empirically estimate $\psi=1.3$.

The goal is to terminate the search in success by revealing $v\%$ of the target, while minimizing the total action cost $n_p + \psi n_s$, where $n_p$ and $n_s$ are the number of pushing and suction actions respectively.
\begin{figure*}[ht!]
\vspace{4pt}
    \centering
    \includegraphics[width=\linewidth]{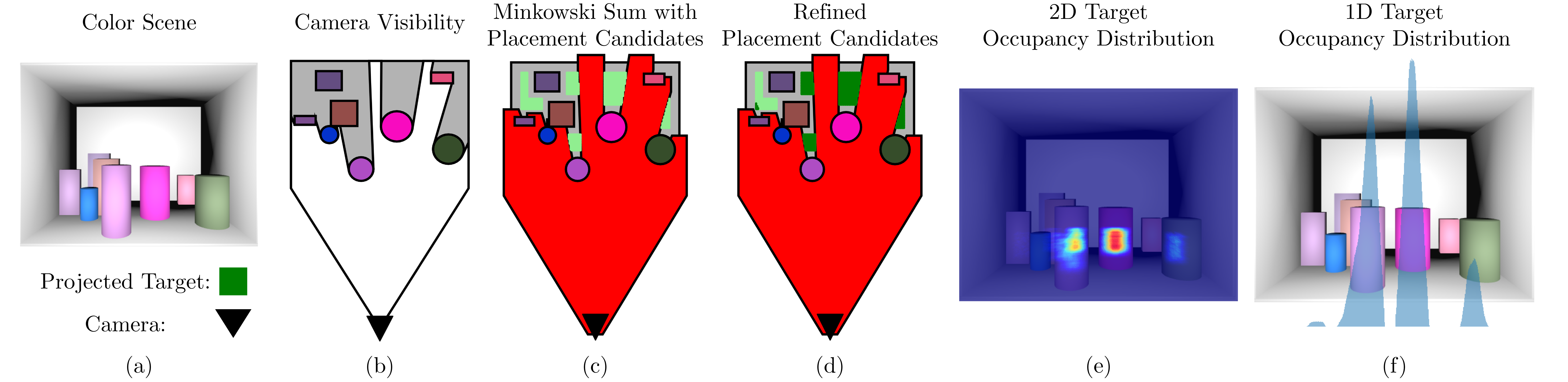}
    \caption{SLAX-RAY Simulation Pipeline: (a) A random set of occluding objects are placed in stable poses such that they do not collide with each other or the shelf. (b) We find the visibility polygon given a projected scene from the camera perspective. (c) We calculate the Minkowski sum of the visibility polygon and the projected target object, which yields a set of candidate target object placements (light green). (d) By casting 3D rays from the camera to all non-colliding candidate target placements, we find the set of all fully hidden target object placements (dark green). (e-f) We project the target object points from each hidden placement into the camera to get a 2D distribution over target locations in image space, then project this 2D distribution to a 1D distribution along the image x-axis.}
    \label{fig:dataset_generation}
    \vspace{-4pt}
\end{figure*}

\section{Bluction Tool}
To facilitate lateral-access mechanical search, we propose a \bluc tool.
A combination of a blade and a suction cup, the bluction tool attaches to the robot wrist, enabling it to perform pushing actions with the blade, and suction actions with the suction cup, as shown in Fig.~\ref{fig:bluction_splash}. Our specific design for the Fetch robot includes an 0.40\,m aluminum vacuum tube with a diameter of 0.03\,m that allows the robot to reach deep into a shelf without losing pressure while avoiding undesirable collisions between the robot wrist and the environment. The blade has a width of 0.065\,m and a height of 0.075\,m. The suction cup at the end has a diameter of 0.03\,m. The bluction tool includes a manually-adjustable camera mount, and allows for autonomous rotation to facilitate camera visibility. Attached to the camera mount is a RealSense L515 LiDAR Camera.

The bluction tool enables a wider set of actions than considered by suction-only~\cite{danielczuk2019mechanical} and push-only~\cite{huang2020mechanical,kurenkov2020visuomotor} mechanical search policies. To push an object with the blade, the robot inserts the flat face of the blade next to the object and then pushes along the blade's surface normal.  To perform a suction action, the robot first rotates the pushing blade upwards to avoid collision with the shelf, before making contact between the suction cup and object.  To take an image of the shelf, the robot rotates the shaft to avoid occlusion of the view by the blade.
\section{Methods}
\label{sec:methods}
In previous work, we presented X-RAY (maXimize Reduction in support Area of occupancY distribution)~\cite{danielczuk2020x} and LAX-RAY (Lateral Access X-RAY)~\cite{huang2020mechanical}, which introduce mechanical search policies that attempt to maximally reduce either support area or entropy of an estimated target occupancy distribution. The target occupancy distribution encodes target object likelihoods within an RGBD image observation, and is estimated using a deep neural network trained on a dataset of simulated depth images and target object segmentation masks with corresponding ground-truth occupancy distributions. To perform mechanical search in a lateral-access environment with the bluction tool, we propose Suction LAX-RAY (SLAX-RAY). SLAX-RAY improves the full LAX-RAY pipeline, including scene generation, perception system, and search policies.
First, SLAX-RAY simulates scenes that resemble cluttered shelf scenarios it might encounter. SLAX-RAY then uses these scenes to train a perception system that predicts the occupancy distribution for a target object. The search policies use this occupancy distribution to determine which action to execute.

To assess the effects of state uncertainty, object arrangement, and available actions on the complexity of the lateral-access mechanical search problem, we also propose an oracle policy that uses full state knowledge to upper-bound policy performance given a scene and available action set.

\subsection{Lateral-Access Simulation}
The SLAX-RAY simulator uses scene generation both to train the perception system and to evaluate policies in simulation experiments. The intent is to generate a set of scenes with random clutter on a shelf, and to render depth images and their corresponding target object occupancy distributions, as in prior work~\cite{huang2020mechanical}.
SLAX-RAY improves on prior work by %
increasing the computational efficiency of the scene generation process and guaranteeing that target objects are completely occluded in all generated scenes.
This new method uses a hybrid approach that relies on 2D visibility calculation, Minkowski sums, and 3D ray-casting. 

To generate a dataset, the scene generator first samples $N \in \irange{2}{12}$ 3D cuboids or cylinders from 6\,cm to 20\,cm in width, depth, and height. It places these objects in one of their stable poses on the shelf iteratively, drawing their 2D positions uniformly at random and rejecting positions that are in collision with the shelf or previously placed objects. Next, it projects the objects to the support surface of the shelf and calculates the 2D visibility of the scene from the perspective of the camera using the CGAL~\cite{cgal:eb-21b} implementation of~\citet{bungiu2014efficient} (Fig.~\ref{fig:dataset_generation}(b)). Given that it knows the projected target object, it calculates a Minkowski sum of the visibility polygon and the convex hull of the target's projected vertices (the red shaded area in Fig.~\ref{fig:dataset_generation}(c)). By sampling in the remaining 2D positions of the shelf, it generates a set of candidate target placements (light green area in Fig.~\ref{fig:dataset_generation}(c)) and refines them by casting rays from the camera to a set of 3D points sampled from the target mesh and transformed to each location; if any ray from the camera to the transformed point does not intersect with the scene, the target object would be visible at that location. Fig.~\ref{fig:dataset_generation}(d) shows the final set of target locations in dark green. Given this set of 2D target locations and a set of sampled mesh points that are transformed to each possible target location, we can directly generate a 2D distribution over possible target object locations in camera frame by projecting the set of target points into the camera (Fig.~\ref{fig:dataset_generation}(e)). Then, by further projecting this 2D distribution to the image $x$-axis, we can reduce it to a 1D distribution (Fig.~\ref{fig:dataset_generation}(f)). In comparison to prior work, which exhaustively transformed the object across a grid of 3D locations in the shelf, this method is over 300\% faster, with generation of a SLAX-RAY training dataset of 100,000 scenes taking only 4 hours on an Ubuntu 20.04 machine with an Intel i7 12-core processor and NVIDIA Titan X GPU.

\subsection{SLAX-RAY Perception System}
Based on the neural network architecture and training pipeline in~\citet{danielczuk2020x}, the SLAX-RAY perception system provides the search policy with information on where the target object may be hidden by computing a target occupancy distribution from a depth image. It also segments the scene into object instances using the input depth image.
The neural network is trained on a dataset of 100,000 depth image and occupancy distribution pairs for three different target object aspect ratios (target objects of size $0.06 m \times 0.06 m \times 0.03 m$, $0.06 m \times 0.06 m \times 0.06 m$, and $0.06 m \times 0.06 m \times 0.12 m$). Each training example contains a single target object and loss for each predicted distribution is only enforced on the network head that outputs the prediction of the aspect ratio corresponding to the target. By training in this way, the network learns to predict a target occupancy distribution for a set of target aspect ratios for each input image. Training takes approximately 18 hours on a Titan X GPU.
At run time, the network takes in a depth image from the physical system and we take the output that corresponds to the aspect ratio of the known target object. SLAX-RAY projects the 2D image-space target occupancy distribution into a 1D distribution along the image $x$-axis, using the assumption that objects rest on the shelf (Sec.~\ref{sec:problemstatement}).  Specifically, the 1D distribution is $P_t(x) = \sum_{y=0}^{h-1} p_t(x, y)$, where $p_t(x,y)$ is the 2D occupancy distribution at pixel location $(x, y)$.

\begin{figure*}
\vspace{4pt}
    \centering
    \includegraphics[width=\linewidth]{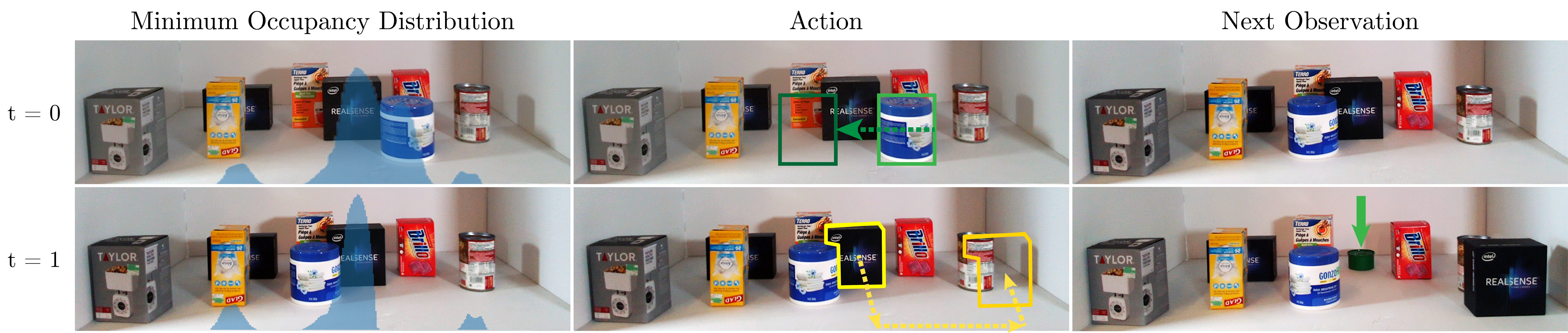}
    \caption{At each timestep, Bluction-DAR computes the minimum of the predicted target occupancy distribution and the minimum target occupancy distribution from the previous timestep. It executes the pushing action (top, green) or suction action (bottom, yellow) that maximally reduces the support area of the distribution. The dotted arrows represent the motion of the end-effector from the lighter colored box to the darker color. Suction actions are advantageous when moving objects that are very close or touching, as in the bottom row.}
    \label{fig:bluction_dar_rollout}
    \vspace{-8pt}
\end{figure*}

\subsection{SLAX-RAY Mechanical Search Policy} \label{subsec:policies}
We propose a novel mechanical search policy: \emph{Bluction-DAR} (Distribution Area Reduction). Bluction-DAR takes actions at each timestep that attempt to maximally reduce the target occupancy distribution until $v\%$ of the target is revealed.
The policy operates on the full action space of pushing $\mathcal{A}_p$ and suction $\mathcal{A}_s$ actions using the \bluc{} tool.
The policy considers the suction action space $\mathcal{A}_s$ to be all combinations of a collision-free grasp and a corresponding collision-free placement, which expands the total action space of the problem beyond the pushing actions previously considered and allows for revealing the target object in scenarios where the blade cannot be inserted between any pair of objects. We restrict suction actions to the piecewise linear action described in Sec.~\ref{sec:problemstatement} to prevent collisions between objects in the shelf. For pushing actions, an object can only be pushed next to the closest object on its left or right such that it does not collide with other objects in the shelf or result in simultaneous pushing of multiple objects, as stated in Sec.~\ref{sec:problemstatement}.

To track search progress and help inform subsequent actions, as in \citet{huang2020mechanical}, we encode the history of previous observations via the minimum of the current 1D predicted distribution and the previous encoding: $P'_t(x) = \min \left\lbrace P_t(x), P'_{t-1}(x)\right\rbrace$. For the first timestep, we set $P'_0(x) = P_0(x)$. We define the support of the occupancy distribution $r_t = |\{ x : P'_t(x) > 0 \} |$. We further define the reduction of support as 
\[\Delta r_t = r_t - r_{t+1},\]
where $r_{t+1}$ is calculated after applying the action. Note that if an action increases the support of the distribution, $\Delta r_t$ may be negative.
We measure $\Delta r_t$ for each action candidate, weighting suction actions by $\psi$ to account for the action cost ratio defined in Sec.~\ref{sec:problemstatement}. Then, for each suction action, $\Delta r^{s}_t = \psi \Delta r_t$, while the reduction of support for the pushing action is kept the same: $\Delta r^{p}_t = \Delta r_t$.
Bluction-DAR returns the list of actions sorted by decreasing $\Delta r^{(\cdot)}_t$, and execute the first kinematically feasible action from the list.

\subsection{Oracle Policies}
\label{ssec:oracle}
We quantify the difficulty of each generated shelf environment and lower-bound policy performance for a given action set by measuring the number of actions taken by oracle policies that use full-state information to reveal the target object. Although policy success is defined in Sec.~\ref{sec:problemstatement} as target object visibility above $v$, the oracle policies search for a series of actions that fully reveals the target. Since all objects are assumed to be cylinders or cuboids, clearing a visibility triangle (from the camera to the front face of the target object in a projected overhead view 
is equivalent to solving the 3D visibility problem. Therefore, for computational efficiency, the oracle policies project the 3D shelf scene into a 2D overhead representation and evaluate actions using the projected representation. 
While oracle policies have full-state knowledge, their action sets are equivalent to the non-oracle policies (e.g., they will not push unreachable objects).

We define two versions of the oracle policy:
\textbf{Oracle-P} (pushing actions only), and \textbf{Oracle-P+S} (pushing and suction actions). Given the visibility triangle to be cleared, the policy exhaustively generates all valid action trees (capped at a maximum of $H$ actions) that would result in this area being free of objects. The action trees are sorted by the number of distinct actions and the policy executes the action tree with the fewest actions. Suction actions are weighted by $\psi$.

\section{Experiments} \label{sec:experiments}
We evaluate SLAX-RAY in both simulated and physical shelf environments. In simulation, the shelf is $0.60$ m wide by $0.60$ m high by $0.60$ m deep. In physical environments, the shelf is $0.80$ m wide by $0.50$ m high by $0.50$ m deep.

\newcommand{\SR}{Success Rate}
\newcommand{\SA}{Median Steps (IQR)}

\newcommand{\sr}[1]{\ifthenelse{\equal{#1}{*}}{\srStar}{\srNoStar{#1}}}
\newcommand{\srStar}[1]{\textbf{\srNoStar{#1}}}
\newcommand{\srNoStar}[1]{#1\,\%}

\newcommand{\sa}[1]{\ifthenelse{\equal{#1}{*}}{\saStar}{\saNoStar{#1}}}
\newcommand{\saStar}[3]{\textbf{\saNoStar{#1}{#2}{#3}}}
\newcommand{\saNoStar}[3]{#1 (#2 -- #3)}

\begin{table*}[t]
\vspace{4pt}
\centering
\begin{tabular}{@{}ccccccc@{}}\toprule
& & 
    \multicolumn{2}{c}{Oracle}  &
    \multicolumn{2}{c}{\citet{huang2020mechanical}} &
    Bluction \\
\cmidrule(lr){3-4} \cmidrule(lr){5-6} \cmidrule(l){7-7} 
No. & Metric & Oracle-P & Oracle-P+S & DAR & DER-3 & Bluction-DAR   \\
\midrule
\multirow{2}*{4} %
  & \SR & \sr{100}        & \sr{100}        & \sr{77}         & \sr{83}         & \sr*{96}   \\
  & \SA & \sa{1.5}{1.0}{2.0} & \sa{1.5}{1.0}{2.0} & \sa{2.0}{1.0}{3.0} & \sa{2.0}{1.0}{3.0} & \sa{2.0}{1.0}{2.5}  \\
\cmidrule{1-7}
\multirow{2}*{6} %
  & \SR & \sr{100}   & \sr{100} & \sr{68}         & \sr{72}        & \sr*{92}       \\
  & \SA & \sa{3.0}{2.0}{3.0} & \sa{2.0}{2.0}{3.0} & \sa{3.0}{1.8}{5.0} & \sa{3.0}{1.0}{4.3} & \sa{2.0}{2.0}{4.0}\\
\cmidrule{1-7}
\multirow{2}*{8} %
  & \SR & \sr{99}  & \sr{100} & \sr{45}         & \sr{50}         & \sr*{88}  \\
  & \SA & \sa{3.0}{2.0}{4.0} & \sa{2.5}{2.0}{3.0} & \sa{3.0}{1.0}{4.0} & \sa{3.0}{2.0}{4.0} & \sa{3.0}{2.0}{5.0}  \\
\cmidrule{1-7}
\multirow{2}*{10} %
  & \SR & \sr{94} & \sr{100}     & \sr{24} & \sr{34}   & \sr*{67}    \\
  & \SA & \sa{4.0}{3.0}{6.0} & \sa{3.0}{2.0}{4.0} & \sa{4.5}{2.0}{6.0}  & \sa{3.0}{2.0}{5.0} & \sa{6.0}{3.0}{8.0} \\
\cmidrule{1-7}
\multirow{2}*{All} %
  & \SR & \sr{98}     & \sr{100}   & \sr{54}   & \sr{60}   & \sr*{86}     \\
  & \SA & \sa{3.0}{2.0}{4.0} & \sa{2.0}{1.0}{3.0} & \sa{2.0}{1.0}{4.0} & \sa{3.0}{1.0}{4.0} & \sa{3.0}{2.0}{4.0} \\
\bottomrule
\end{tabular}
\caption{Simulated policy rollouts: 100 trials were executed for 4, 6, 8, and 10 occluding objects; in total, 2000 trials. For each, we show the \textbf{Success Rate} followed by the \textbf{Median} and \textbf{Interquartile Range (IQR)} for the number of steps taken to reveal the target. Results suggest that suction actions allow the Bluction-DAR policy to succeed between $13\%$ and $38\%$ more often in densely cluttered environments than the best-performing pushing baselines. }
\label{tab:results}
\end{table*}

\subsection{Simulation Experiments}
The simulated experiments use the First Order Shelf Simulator (FOSS) from \citet{huang2020mechanical}. We generate 400 total environments with 4, 6, 8, and 10 occluding objects. For these experiments, we set the maximum number of steps to $H = 2n$, where $n$ is the number of objects. 

We evaluate Bluction-DAR on the set of 400 shelves along with two baselines from previous work: \textbf{DAR} (Distribution Area Reduction) and \textbf{DER-3} (Distribution Entropy Reduction over 3 steps) are lateral-access mechanical search policies based on target occupancy distribution predictions~\cite{huang2020mechanical}. DAR is an ablation of Bluction-DAR that greedily reduces the target object distribution support area and DER-3 is a policy with 3-step lookahead that takes the action with least predicted entropy after the 3 steps. Both DAR and DER-3 only use pushing actions. For these experiments, we use the target object of size $0.06 m \times 0.06 m \times 0.06 m$ and set the target visibility threshold to $v=80\%$. 

Table~\ref{tab:results} shows the results. Bluction-DAR succeeds in revealing the target more than baselines across all scenes, with a $13\%$ improvement over the best-performing pushing-only baseline for 4 object scenes and a $33\%$ improvement over the best-performing baseline for 10 object scenes. Bluction-DAR also maintains success rates of at least $67\%$ even in 10 object scenes, suggesting it scales better to more dense arrangements than baselines. Bluction-DAR gains a considerable advantage in denser scenes because of the additional object mobility. Since objects may not be pushed behind other objects (as to not risk collisions with obscured objects), pushing policies are extremely limited in scenes with many objects. Neighboring objects can block large parts of the scene and it may be difficult or impossible to align the small window of visibility with the target object. Suction actions enable the movement of occluders without having to clear a substantial pushing path. As compared to previous work~\cite{huang2020mechanical}, DAR and DER-3 success rates are lower due to the relaxation of the assumption that objects must be separated from each other and the shelf walls by at least the blade width.

The performance of the oracle policies in Table~\ref{tab:results} suggests that while pushing actions can reveal the target object reliably in less cluttered shelves of 4, 6, or 8 occluding objects, as shelves become more densely packed, the upper bound of pushing policies begins to drop. However, when using both pushing and suction actions, it remains possible to succeed in $100\%$ of environments. These experiments further explain the performance gain of Bluction-DAR over pushing-only baselines and suggest that the bluction tool and associated new actions may be necessary to reveal target objects in increasingly dense scenes. The Oracle-P+S policy outperforms Bluction-DAR by $12\%$ and $33\%$ in scenes with 8 and 10 objects, respectively, which suggests that the Bluction-DAR policy can be further improved to approach success rates over $90\%$ even in denser scenes.

\subsection{Physical Experiments}
We evaluate SLAX-RAY and the DER-3 policy (our best-performing baseline in the simulation experiments) in 9 physical shelf environments with 8, 10, and 12 objects like the one shown in Fig.~\ref{fig:bluction_splash}. We randomly place the household kitchen and bathroom objects with an occluded target of size $0.06m \times 0.06m \times 0.03m$ on a shelf and execute the pushing and suction actions using a Fetch robot with the \bluc tool. We use the Intel RealSense LiDAR Camera L515 for the RGBD observations. The results in Table~\ref{tab:phys_results} suggest that Bluction-DAR performs consistently better than DER-3 across these scenes. A benefit of the suction action is that Bluction-DAR is more robust to the perception noise. Specifically, for 3 out of 9 trials, DER-3 fails to find the available actions when the segmentation mask is narrower than the actual object since it would produce a false collision detection. Bluction-DAR instead is still able to find available actions via suction actions. Bluction-DAR is also more robust to manipulation or trajectory planning deviations. When the robot accidentally pushes objects against other objects or to the wall, Bluction-DAR is able to continue searching using suction actions. Bluction-DAR is also able to find the target in scenes which is unsolvable for pushing only policies such as the scenes where the target is hidden behind two objects with no large gap in between for blade insertion.
\begin{table}[t]
\centering
\begin{tabular}{@{}cccc@{}}\toprule
No. & Metric & DER-3 & Bluction-DAR  \\
\midrule
\multirow{2}*{8} %
  & Successes & 0/3         & \textbf{3/3}  \\
  & No. Steps & ( - ) & 3, 4, 4 \\
\cmidrule{1-4}
\multirow{2}*{10} %
  & Successes & 2/3 & \textbf{3/3}   \\
  & No. Steps & 7, 4  & 4, 5, 4 \\
\cmidrule{1-4}
\multirow{2}*{12} %
  & Successes &1/3 & \textbf{3/3}   \\
  & No. Steps & 4 & 6, 7, 7 \\
\cmidrule{1-4}
\multirow{2}*{All} %
  & \SR & \sr{33}   & \sr*{100}      \\
  & \SA & \sa{4.0}{4.0}{5.5} & \sa{4.0}{4.0}{6.0}  \\
\bottomrule
\end{tabular}
\caption{
Physical experiment results. We evaluate DER-3 and Bluction-DAR across 3 scenes with 8, 10, and 12 objects each. Bluction-DAR outperforms DER-3 in physical scenes consistently.}
\label{tab:phys_results}
\vspace{-4pt}
\end{table}

\section{Conclusion and Future Work} \label{sec:discussion}
This work presents \AlgName{}, a lateral access mechanical search pipeline. \AlgName{} uses the bluction tool, a novel perception algorithm and a new search policy Bluction-DAR that uses both pushing and suction actions to reveal a target object in a shelf environment.
Experiments in simulated and physical settings demonstrate that the increased action set improves success rate by $26\%$ and $67\%$, respectively, over pushing-only baselines. In future work, we will explore actions that move multiple objects simultaneously, plan collisions between objects to ``nudge" occluders out of the way, and explore scenes with stacked objects. We will also explore improvement on SLAX-RAY searching policies by incorporating future predictions.
\section*{Acknowledgements}
\footnotesize
This research was performed at the AUTOLAB at UC Berkeley in affiliation with the Berkeley AI Research (BAIR) Lab, and the CITRIS ``People and Robots'' (CPAR) Initiative. The authors were supported in part by donations from Google, Siemens, Toyota Research Institute, Autodesk, Honda, Intel, and Hewlett-Packard. This material is based upon work supported by the National Science Foundation Graduate Research Fellowship Program under Grant No. DGE 1752814. Any opinions, findings, and conclusions or recommendations expressed in this material are those of the authors and do not necessarily reflect the views of the sponsors. We thank our colleagues who provided helpful feedback and suggestions, in particular Lawrence Chen, Vishal Satish, and Kate Sanders.

\renewcommand*{\bibfont}{\footnotesize}
\printbibliography
\end{document}